\documentclass[10pt, a4paper]{article}
\usepackage{lrec}
\usepackage{multibib}
\newcites{languageresource}{Language Resources}
\usepackage{graphicx}
\usepackage{tabularx}
\usepackage{soul}
\usepackage{url}
\usepackage{amsmath}

\usepackage{epstopdf}
\usepackage[utf8]{inputenc}

\usepackage{hyperref}
\usepackage{xstring}
\usepackage{xcolor}

\usepackage{color}

 \def\paperDraft{}
 \ifdefined\paperDraft
  
   \def\spcomment#1{{\color{blue}[Senja: \textit{#1}]}}
\else
 \def\spcomment#1{}
 
\fi
\title{Leveraging Contextual Embeddings for Detecting Diachronic Semantic Shift \\ \vspace*{.5\baselineskip}}

\name{Matej Martinc, Petra Kralj Novak, Senja Pollak}

\address{Jozef Stefan Institute \\
         Ljubljana, Slovenia \\
         \{matej.martinc, petra.kralj.novak, senja.pollak\}@ijs.si\\}

\abstract{
We propose a new method that leverages contextual embeddings for the task of diachronic semantic shift detection by generating time specific word representations from BERT embeddings. The results of our experiments in the domain specific LiverpoolFC corpus suggest that the proposed method has performance comparable to the current state-of-the-art without requiring any time consuming domain adaptation on large corpora. The results on the newly created Brexit news corpus suggest that the method can be successfully used for the detection of a short-term yearly semantic shift. And lastly, the model also shows promising results in a multilingual settings, where the task was to detect differences and similarities between diachronic semantic shifts in different languages.\\ \newline  }

\begin{document}

\maketitleabstract

\section{Introduction}

While language is many times mistakenly perceived as a stable, unchanging structure, it is in fact constantly evolving and adapting to the needs of its users. It is a well researched fact that some words and phrases can change their meaning completely in a longer period of time. The word \textit{gay}, which was a synonym for cheerful until the 2\textsuperscript{nd} half of the 20\textsuperscript{th} century, is just one of the examples found in the literature. On the other hand, we are just recently beginning to research and measure more subtle semantic changes that occur in much shorter time periods. These changes reflect a sudden change in language use due to changes in the political and cultural sphere or due to the localization of language use in somewhat closed communities.  

The study of how word meanings change in time has a long tradition \cite{bloomfield1933language} but it has only recently saw a surge in popularity and quantity of research due to recent advances in modelling semantic relations with word embeddings \cite{mikolov2013efficient} and increased availability of textual resources. The current state-of-the-art in modelling semantic relations are contextual embeddings \cite{devlin2018bert,peters2018deep}, where the idea is to generate a different vector for each context a word appears in, i.e., for each specific word occurrence. This solves the problems with word polysemy and employing this type of embeddings has managed to improve the state-of-the-art on a number of natural language understanding tasks. However, contextual embeddings have not yet been widely employed in the discovery of diachronic semantic shifts.
 
In this study, we present a novel method that relies on contextual embeddings to generate time specific word representations that can be leveraged for the purpose of diachronic semantic shift detection \footnote{Code is available at \url{https://gitlab.com/matej.martinc/semantic_shift_detection}.}. We also show that the proposed approach has the following advantages over existing state-of-the-art methods:
 
\begin{itemize}
    \item It shows comparable performance to the previous state-of-the-art in detecting a short-term semantic shift without requiring any time consuming domain adaptation on a very large corpus that was employed in previous studies.
    \item It enables the detection and comparison of semantic shifts in a multilingual setting, which is something that has never been automatically done before and will facilitate the research of differences and similarities of how word meanings change in different languages and cultures.
\end{itemize}
 
The paper is structured as follows. We address the related work on diachronic semantic shift detection in Section \ref{sec:relatedWork} We describe the methodology and corpora used in our research in Section \ref{sec:methodology} The conducted experiments and results are presented in Section \ref{sec:experiments} Conclusions and directions for further work are presented in Section~\ref{sec:conclusion}

\section{Related Work}
\label{sec:relatedWork}

If we take a look at a research on diachronic semantic shift, we can identify two distinct trends: (1) a shift from raw word frequency methods to methods that leverage dense word representations, and (2) a shift from long-term semantic shifts (spanning decades or even centuries) to short-term shifts spanning years at most.

Earlier studies \cite{juola2003time,hilpert2008assessing} in detecting semantic shift and linguistic change used raw word frequency methods for detecting semantic shift and linguistic change. They are being replaced by methods that leverage dense word representations. The study by \newcite{kim2014temporal} was arguably the first that employed word embeddings, or more specifically, the Continuous Skipgram model proposed by \newcite{mikolov2013efficient}, while the first research to show that these methods can outperform frequency based methods by a large margin was conducted by \newcite{kulkarni2015statistically}. 

In the latter method, separate word embedding models are trained for each of the time intervals. Since embedding algorithms are inherently stochastic and the resulting embedding sets are invariant under rotation, vectors from these models are not directly comparable and need to be aligned in a common space \cite{kutuzov2018diachronic}.
To solve this problem, \newcite{kulkarni2015statistically} first suggested a simple linear transformation for projecting embeddings into a common space. \newcite{zhang2016past} improved this approach by proposing the use of an additional set of nearest neighbour words from different models that could be used as anchors for alignment. Another approach was devised by \newcite{eger2017linearity}, who proposed second-order embeddings (i.e., embeddings of word similarities) for model alignment and it was \newcite{hamilton2016cultural} that showed that these two methods can compliment each other.

Since imperfect aligning can negatively affect semantic shift detection, the newest methods try to avoid it altogether. \newcite{rosenfeld2018deep} presented an approach, where the embedding model is trained on word and time representations, treating the same words in different time periods as different tokens. Another solution to avoid alignment is the incremental model fine-tuning, where the model is first trained on the first time period and saved. The weights of this initial model are used for the initialization of the model trained on the next successive time period. The described step of incremental weight initialization is repeated until the models for all time periods are trained. This procedure was first proposed by \newcite{kim2014temporal} and made more efficient by \newcite{peng2017incrementally}, who suggested to replace the softmax function for the Continuous bag-of-word and Continuous skipgram models with a more efficient hierarchical softmax, and by \newcite{kaji2017incremental}, who proposed an incremental extension for negative sampling.

Recently, a new type of embeddings called contextual embeddings have been introduced.  ELMo (Embeddings from Language Models) by \newcite{peters2018deep} and BERT (Bidirectional Encoder Representations from Transformers) by \newcite{devlin2018bert} are the most prominent representatives of this type of contextual embeddings. In this type of embeddings, a different vector is generated for each context a word appears in. These new contextual embeddings solve the problems with word polysemy but have not been used widely in the studies concerning temporal semantic shifts. The only two temporal semantic shift studies we are aware off, that used contextual BERT embeddings, are reported in \newcite{hu-etal-2019-diachronic} and \newcite{giulianellilexical}. 

In the study by \newcite{hu-etal-2019-diachronic}, contextualised BERT embeddings were leveraged to learn a representation for each word sense in a set of polysemic words. Initially, BERT is applied to a diachronic corpus to extract embeddings for tokens that closely match the predefined senses of a specific word. After that, a word sense distribution is computed at each successive time slice. By comparing these distributions, one is able to inspect the evolution of senses for each target word.

In the study by \newcite{giulianellilexical}, word meaning is considered as ``inherently under determined and contingently modulated in situated language use'', meaning that each appearance of a word represents a different word usage. The main idea of the study is to determine how word usages vary through time. First, they fine-tune the BERT model on the entire corpus for domain adaptation and after that they perform diachronic fine-tuning, using the incremental training approach proposed by \newcite{kim2014temporal}. After that, the word usages for each time period are clustered with the K-means clustering algorithm and the resulting clusters of different word usages are compared in order to determine how much the word usage changes through time.

The second trend in diahronic semantic change research is a slow shift of focus from researching long-term semantic shifts spanning decades or even centuries to short-term shifts spanning years at most \cite{kutuzov2018diachronic}. For example, a somewhat older research by \newcite{sagi2011tracing} studied differences in the use of English spanning centuries by using the Helsinki corpus \cite{rissanen1993helsinki}. The trend of researching long-term shifts continued with \newcite{eger2017linearity} and \newcite{hamilton2016diachronic}, who both used the Corpus of Historical American (COHA)\footnote{\url{http://corpus.byu.edu/coha}}. In order to test if existing methods could be applied to detect short-term semantic changes in language, newer research focuses more on tracing short-term socio-cultural semantic shift. \newcite{kim2014temporal} analyzed yearly changes of words in the Google Books Ngram corpus and \newcite{kulkarni2015statistically} analyzed Amazon Movie Reviews, where spans were one year long, and Tweets, where change was measured in months. The most recent exploration of meaning shift over short periods of time that we are aware of, was conducted by \newcite{del2019short}, who measured changes of word meaning in online Reddit communities by employing the incremental fine-tuning approach proposed by \newcite{kim2014temporal}. 

\section{Methodology}
\label{sec:methodology}

In this section, we present the methodology of the proposed approach by explaining how we obtain time period specific word representations, on which corpora the experiments are conducted, and how we evaluate the approach.

\subsection{Time specific word representations}
\label{sec:time_rep}

Given a set of corpora containing documents from different time periods, we develop a method for locating words with different meaning in different time periods and for quantifying these meaning changes. Our methodology is similar to the approach proposed by \newcite{rosenfeld2018deep} since we both construct a time period specific word representation that represents a semantic meaning of a word in a distinct time period.

In the first step, we fine-tune a pretrained BERT language model for domain adaptation on each corpus presented in Section \ref{sec:corpora} Note that we do not conduct any diachronic fine-tuning, therefore our fine-tuning approach differs from the approach presented in \newcite{giulianellilexical}, where BERT contextual embeddings were also used, and also from other approaches from the related work that employ the incremental fine-tuning approach \cite{kim2014temporal,del2019short}. The reason behind this lies in the contextual nature of embeddings generated by the BERT model, which are by definition dependent on the time-specific context and therefore, in our opinion, do not require diachronic (time-specific) fine-tuning. We use the English BERT-base-uncased model with 12 attention layers and a hidden layer size of 768 for experiments on the English corpora, and the multilingual BERT-base-cased model for multilingual experiments\footnote{Although recently a variety of novel transformer language models emerged, some of them outperforming BERT \cite{yang2019xlnet,sun2019ernie}, BERT was chosen in this research due to the availability of the pretrained multilingual model which among other languages also supports Slovenian.}. Only one model is used for generating the time period specific word representations in the multilingual setting and not two: one for each language in our experiments, English and Slovenian. We opted for this method in order to generate word representations for both languages that do not need to be aligned in a common vector space and are directly comparable. We only conduct light text preprocessing on the LiverpoolFC corpus, where we remove URLs.

In the next step, we generate time specific representations of words. Each corpus is split into predefined time periods and a set of time specific subcorpora is created for each corpus. The documents from each of the time specific subcorpora are split into sequences of byte-pair encoding tokens \cite{kudo2018sentencepiece} of a maximum length of 256 tokens and fed into the fine-tuned BERT model. For each of these sequences of length $n$, we create a sequence embedding by summing the last four encoder output layers. The resulting sequence embedding of size $n$ times \textit{embeddings size} represents a concatenation of contextual embeddings for the $n$ tokens in the input sequence. By chopping it into $n$ pieces, we acquire a representation, i.e., a contextual token embedding, for each word usage in the corpus. Note that these representations vary according to the context in which the token appears, meaning that the same word has a different representation in each specific context (sequence). Finally, the resulting embeddings are aggregated on the token level (i.e., for every token in the corpus vocabulary, we create a list of all their contextual embeddings) and averaged, in order to get a time specific representation for each token in each time period. 

Last, we quantitatively estimate the semantic shift of each target word in the period between two time specific representations by measuring the cosine distance between two time specific representations of the same token. This differs from the approach proposed by \newcite{giulianellilexical}, where clustering was used as an aggregation method and than Jensen-Shannon divergence was measured, a measure of similarity between probability distributions, to quantify changes between word usages in different time periods.  

Another thing to note is that for the experiments on the Brexit news corpus (see Section \ref{sec:corpora}), we conduct the same averaging procedure on the entire corpus (not just on the time specific subcorpus) in order to get a general (not just time specific) representation for each token in the corpus. These general representations of words are used to find the 50 most similar words to the word \textit{Brexit} (see Section \ref{sec:brexit} for further details).

Since the byte-pair input encoding scheme \cite{kudo2018sentencepiece} employed by the BERT model does not necessarily generate tokens that correspond to words but rather generate tokens that can sometimes correspond to subparts of words, we also propose the following \textit{on the fly} reconstruction mechanism that allows us to get word representations from byte pair tokens. If a word is split into more than one byte pair tokens, we take an embedding for each byte pair token constituting a word and build a word embedding by averaging these byte pair tokens. The resulting average is used as a context specific word representation.

\subsection{Corpora}
\label{sec:corpora}

We used three corpora in our experiments, all of them covering short time periods of eight years or less. The statistics about the datasets are presented in Table \ref{tbl:datasets}. 

\subsubsection{LiverpoolFC}
\label{sec:liverpool}

The LiverpoolFC corpus is used to compare our approach to a recent state-of-the-art approach proposed by \newcite{del2019short}. It contains 8 years of Reddit posts, more specifically the LiverpoolFC subreddit for fans of the English football team. It was created for the task of short-term meaning shift analysis in online communities. The language use in the corpus is specific to a somewhat closed community, which means linguistic innovations are common and non-standard word interpretations are constantly evolving. This makes this corpus very appropriate for testing the models for abrupt semantic shift detection. 

We adopt the same procedure as the original authors and split the corpus into two time spans, the first one covering texts ranging from 2011 until 2013 and the second one containing texts from 2017.

\subsubsection{Brexit news}

We compiled the Brexit news corpus to test the ability of our model to detect relative semantic changes (i.e., how does a specific word, in this case \textit{Brexit}, semantically correlate to other words in different time periods) and to test the method on consecutive yearly periods. The subject of Brexit was chosen due to its extensive news coverage over a longer period of time, which allows us to detect possible correlations between the actual events that occurred in relation to this topic and semantic changes detected by the model. The corpus contains about 36.6 million tokens and consists of news articles (more specifically, their titles and content) about Brexit\footnote{Only articles that contain word \textit{Brexit} in the title were used in the corpus creation.} from the RSS feeds of the following news media outlets: Daily Mail, BBC, Mirror, Telegraph, Independent, Guardian, Express, Metro, Times, Standard and Daily Star and the Sun. The corpus is divided into 5 time periods, the first one covering articles about the Brexit before the referendum that occurred on June 23, 2016. The articles published after the referendum are split into 4 yearly periods. The yearly splits are made on June 24 each year and the most recent time period contains only articles from June 24, 2019 until August 23, 2019. The corpus is unbalanced, with time periods of 2016 and 2018 containing much more articles than other splits due to more intensive news reporting. See Table \ref{tbl:datasets} for details.

\subsubsection{Immigration news}

The Immigration news corpus was compiled to test the ability of the model to detect relative semantic changes in a multilingual setting, something that has to our knowledge never been tried before. The main idea is to detect similarities and differences in semantic changes related to immigration in two distinct countries with different attitudes and historical experiences about this subject. 

The topic of immigration was chosen due to relevance of this topic for media outlets in both countries that were covered, England and Slovenia. The corpus consists of 6,247 English articles and 10,089 Slovenian news articles (more specifically, their titles and content) about immigration\footnote{The corpus contains English articles that contain words \textit{immigration}, \textit{immigrant} or \textit{immigrants} in the title and Slovenian articles that contain Slovenian translations of these words in either title or content.}, is balanced in terms of number of tokens for each language and altogether contains about 12 million tokens. The English and Slovenian documents are combined and shuffled\footnote{Shuffling is performed to avoid the scenario where all English documents would be at the beginning of the corpus and all Slovenian documents at the end, which would negatively affect the language model fine-tuning.} and after that the corpus is divided into 5 yearly periods (split on December 31). The English news articles were gathered from the RSS feeds of the same news media outlets as the news about Brexit, while the Slovenian news articles were gathered from the RSS feeds of the following Slovenian news media outlets: Slovenske novice, 24ur, Dnevnik, Zurnal24, Vecer, Finance and Delo. 

\begin{table}[!h]
\begin{center}
\begin{tabularx}{\columnwidth}{|l|l|X|}

      \hline
      Corpus & Time period & Num. tokens (in millions)\\
      \hline
      LiverpoolFC & 2013 & 8.5 \\
      LiverpoolFC & 2017 & 11.9 \\
      LiverpoolFC & Entire corpus & 20.4 \\
      \hline
      Brexit news & 2011 - 23.6.2016 & 2.6 \\
      Brexit news & 24.6.2016 - 23.6.2017 & 10.3\\
      Brexit news & 24.6.2017 - 23.6.2018 & 6.2\\
      Brexit news & 24.6.2018 - 23.6.2019 & 12.7\\
      Brexit news & 24.6.2019 - 23.8.2019 & 2.4\\
      Brexit news & Entire corpus & 36.6\\
      \hline
      Immigration news & 2015 & 2.2\\
      Immigration news & 2016 & 2.6\\
      Immigration news & 2017 & 2.6\\
      Immigration news & 2018 & 2.6\\
      Immigration news & 2019 & 1.9\\
      Immigration news & Entire corpus & 11.9\\
      \hline
\end{tabularx}
\caption{Corpora statistics.}
\label{tbl:datasets}
 \end{center}
\end{table}

\subsection{Evaluation}
\label{sec:eval}

We evaluate the performance of the proposed approach for semantic shift detection by conducting quantitative and qualitative evaluation.

\subsubsection{Quantitative evaluation}

In order to get a quantitative assessment of the performance of the proposed approach, we leverage a publicly available evaluation set for semantic shift detection on the LiverpoolFC corpus \cite{del2019short}. The evaluation set contains 97 words from the corpus manually annotated with semantic shift labels by the members of the LiverpoolFC subreddit. 26 community members with domain knowledge but no linguistic background were asked to make a binary decision whether the meaning of the word changed between the two time spans (marked as 1) or not (marked as 0) for each of the words in the evaluation set. Each word received on average 8.8 judgements and the average of these judgements is used as a gold standard semantic shift index. 

Positive examples of meaning shift in this evaluation set can be grouped into three classes according to the type of meaning shift. First are metonymic shifts, which are figures of speech, in which a thing or concept is referred to by the name of something associated with it (e.g., the word \textit{F5} that is initially used as a shortcut for refreshing a page and starts to denote any act of refreshing). Second are metaphorical shifts where the original meaning of a word is widened through analogy (e.g., the word \textit{pharaoh} which is the nickname of an Egyptian football player). Lastly, memes are semantic shifts that occur when a word first used in a humorous or sarcastic way prompts a notable change in word's usage on a community scale (e.g., the first part of the player's surname \textit{Van Dijk} is being used in jokes related to shoes' brand \textit{Vans}).

We measure Pearson correlation between the semantic shift index and the model's semantic shift assessment for each of the words in the evaluation set in order to be able to directly compare our approach to the one presented in \newcite{del2019short}, where the same evaluation procedure was employed. As explained in Section \ref{sec:time_rep}, we obtain semantic shift assessments by measuring the cosine distance between two time specific representations of the same token.

\subsubsection{Qualitative evaluation}

For the Brexit and Immigration news corpora, manually labeled evaluation sets are not available, therefore we were not able to quantitatively assess the approach's performance on these two corpora. For this reason, the performance of the model on these two corpora is evaluated indirectly, by measuring how does a specific word of interest semantically correlate to other seed words in a specific time period and how does this correlation vary through time. The cosine distance between the time specific representation of a word of interest and the specific seed word is used as a measure of semantic relatedness. We can evaluate the performance of the model in a qualitative way by exploring if detected differences in semantic relatedness (i.e., relative semantic shifts) are in line with the occurrences of relevant events which affected the news reporting about Brexit and Immigration, and also the findings from the academic studies on these topics. This is possible because topics of Brexit and Immigration have been extensively covered in the news and several qualitative analyses on the subject have been conducted. 

The hypothesis that justifies this type of evaluation comes from structural linguistics and states that word meaning is a relational concept and that words obtain meaning only in relation to their neighbours \cite{matthews2001short}. According to this hypothesis, the change in the word's meaning is therefore expressed by the change in semantic relatedness to other neighbouring words. Neighbouring seed words to which we compare the word of interest for the Brexit news corpus are selected automatically (see Section \ref{sec:brexit} for details) while for the Immigration news corpus, the chosen seed words are concepts representing most common aspects of the discourse about immigration (see Section \ref{sec:immigration} for details).

\section{Experiments}
\label{sec:experiments}

In this section we present details about conducted experiments and results on the LiverpoolFC, Brexit and Immigration corpora.

\subsection{LiverpoolFC}

\begin{figure*}[!h]
\begin{center}
\includegraphics[scale=0.29]{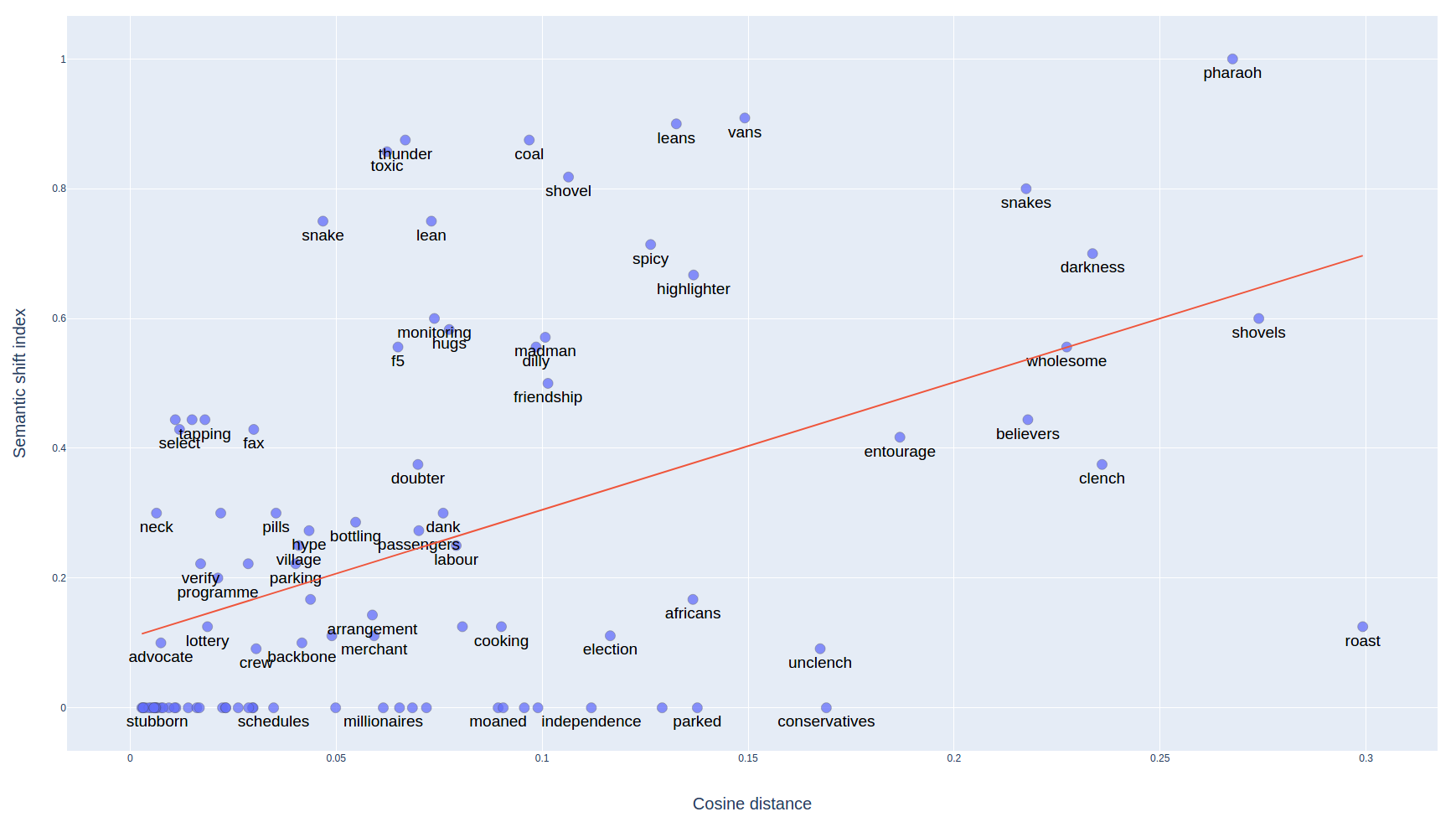} 
\caption{Semantic shift index vs. cosine distance in the LiverpoolF1 evaluation dataset.}
\label{fig.1}
\end{center}
\end{figure*}

In this first experiment, we offer a direct comparison of the proposed method to the state-of-the-art approach proposed by \newcite{del2019short}. In their study, they use a Continuous Skipgram model proposed by \newcite{mikolov2013efficient} and employ the incremental model fine-tuning approach first proposed by \newcite{kim2014temporal}. In the first step, they create a large Reddit corpus (with about 900 million tokens) containing Reddit post from the year 2013 and use it for training the domain specific word embeddings. The embeddings of this initial model are used for the initialization of the model trained on the next successive time period, LiverpoolFC 2013 posts, and finally, the embeddings of the LiverpoolFC 2013 model are used for the initialization of the model trained on the LiverpoolFC 2017 posts. We, on the other hand, do not conduct any additional domain adaptation on a large Reddit corpus and only fine-tune the BERT model on the LiverpoolFC corpus, as already explained in Section \ref{sec:methodology}.

First, we report on the results of the diachronic semantic shift detection for 97 words from the LiverpoolFC corpus that were manually annotated with semantic shift labels by members of the LiverpoolFC subreddit (see Section \ref{sec:liverpool} for more details on the annotation and evaluation procedures). Overall, our proposed approach yields almost identical positive correlation between cosine distance between 2013 and 2017 word representations and semantic shift index as in the research conducted by \newcite{del2019short}. We observe the Pearson correlation of 0.47 (p $<$ 0.001) while the original study reports Pearson correlation of 0.49. 

On the other hand, there are also some important differences between the two methods. Our approach (see Figure \ref{fig.1}) proves to be more conservative when it comes to measuring the semantic shift in terms of cosine distance. In the original approach, the cosine distance of up to 0.6 is measured for some of the words in the corpus, while we only observe the differences in cosine distance of up to 0.3 (for the word \textit{roast}). This conservatism of the model results in less false positive examples (i.e., detected semantic shifts that were not observed by human annotators) compared to the original study, but also results in more false negative examples (i.e., unrecognised semantic shifts that were recognized by human annotators)\footnote{Expressions \textit{false positive} and \textit{false negative} are used here to improve the readability of the paper and should not be interpreted in a narrow context of binary classification.}. An example of a false negative detection by the system proposed by \newcite{del2019short} is the word \textit{lean}. An example of a false positive detection by the system proposed by \newcite{del2019short} that was correctly identified by our system as a word with unchanged semantic context is the word \textit{stubborn}. On the other hand, our system also manages to correctly identify some of the words that changed the most that were misclassified by the system proposed by \newcite{del2019short}. An example of this is the word \textit{Pharaoh}.

There are also some similarities between the two systems. For example, the word \textit{highlighter} is correctly identified as a word that changed meaning by both systems. With the exception of \textit{Pharaoh}, we also notice similar tendencies of both systems to misclassify as false negatives words that fit into the category of so-called metaphorical shifts (i.e., widening of the original meaning of a word through analogy). Examples of these words would be \textit{snake},  \textit{thunder} and \textit{shovel}. One explanation for this misclassification that was offered by \newcite{del2019short} is the fact that many times the metaphoric usage is very similar to the literal one, therefore preventing the model to notice the difference in meaning\footnote{For example, \textit{shovel} is used in a context where the team is seen as a train running through the season, and the fan's job is to contribute in a figurative way by shoving the coal into the train boiler. Therefore, the word \textit{shovel} is used in sentences like \textit{You boys know how to shovel coal.}}.

\subsection{Brexit news}
\label{sec:brexit}

\begin{figure*}[!h]
\begin{center}
\includegraphics[scale=0.6]{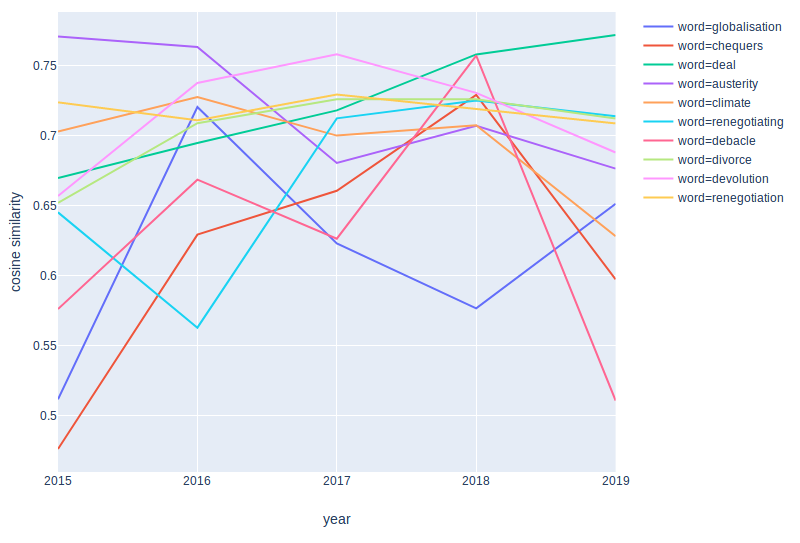} 
\caption{The relative diachronic semantic shift of the word \textit{Brexit} in relation to ten words that changed the most out of 50 closest words to \textit{Brexit} according to the cosine similarity.}
\label{fig.2}
\end{center}
\end{figure*}

Here we asses the performance of the proposed approach for detecting sequential semantic shift of words in short-term yearly periods. More specifically, we explore how time specific seed word representations in different time periods change their semantic relatedness to the time specific word representation of the word \textit{Brexit}. The following procedure is conducted. First, we find 50 words most semantically related to the general non-time specific representation of \textit{Brexit} according to their non-time specific general representations. Since the initial experiments showed that many of the 50 most similar words are in fact derivatives of the word \textit{Brexit} (e.g., \textit{brexitday}, \textit{brexiters}...) and therefore not that relevant for the purpose of this study (as their meaning is fully dependant on the concept from which they derived), we first conduct an additional filtering according to the normalized Levenshtein distance defined as:

\[normLD = 1 - (LD / max(len(w1), len(w2))),\]

where $normLD$ stands for normalized Levenshtein difference, $LD$ for Levenshtein difference, $w1$ is \textit{Brexit} and $w2$ are other words in the corpus. Words for which normalized Levenshtein difference is bigger than 0.5 are discarded and out of the remaining words we extract 50 words most semantically related to \textit{Brexit} according to the cosine similarity. \footnote{The normalized Levenshtein difference treshold of 0.5 and the number of most semantically similar words were chosen empirically.}

Out of these 50 words, we find ten words that changed the most in relation to the time specific representation of the word \textit{Brexit} with the following equation:

\[\mathit{MC} = abs(CS(w1_{2015}, w2_{2015}) - CS(w1_{2019}, w2_{2019}))\]

    where \textit{MC} stands for meaning change, \textit{CS} stands for cosine similarity, $w1_{year}$ is a year specific representation of the word \textit{Brexit} and $w2_{year}$ are year specific representations of words related to \textit{Brexit}. 

\begin{figure*}[!t]
\begin{center}
\includegraphics[scale=0.6]{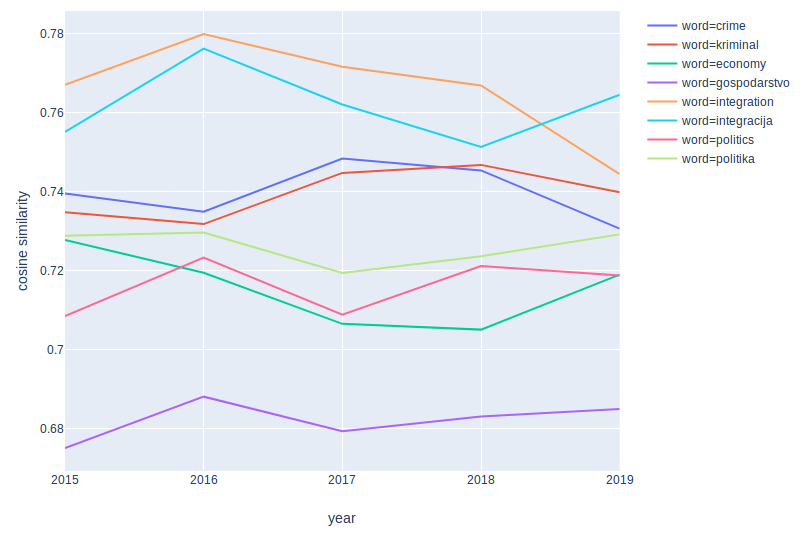} 
\caption{The relative diachronic semantic shift of the word \textit{immigration} in relation to English-Slovenian word pairs crime-kriminal, economy-gospodarstvo, integration-integracija and politics-politika.}
\label{fig.3}
\end{center}
\end{figure*}

The resulting 10 seed words are used to determine the relative diachronic semantic shift of the word \textit{Brexit} as explained in Section \ref{sec:eval} Figure \ref{fig.2} shows the results of the experiments. We can see that the word \textit{deal} is constantly becoming more and more related to \textit{Brexit}, from having a cosine similarity to the word \textit{Brexit} of about 0.67 in 2015 to having a cosine similarity of about 0.77 in 2019. This is consistent with our expectations. The biggest overall difference of about 0.14 in semantic relatedness can be observed for the word \textit{globalisation}, having not been very related to \textit{Brexit} before the referendum in year 2016 (with the cosine similarity of about 0.52) and than becoming very related to the word \textit{Brexit} in a year after the referendum (with cosine similarity of about 0.72). After that, we can observe another drop in similarity in the following two years and then once again a rise in similarity in 2019. This movement could be at least partially explained by the post-referendum debate on whether UK's \textit{Leave} vote could be seen as a vote against globalisation \cite{coyle2016brexit}.

A sudden rise in semantic relatedness between words \textit{Brexit} and \textit{devolution} in years 2016 and 2017 could be explained by a still quite relevant question of how UK's withdrawal from the EU will affect its structures of power and administration \cite{hazell2016brexit}. We can also observe a quite sudden drop in semantic relatedness between the words \textit{Brexit} and \textit{austerity} in year 2017, one year after the referendum. It is possible, that the debate on whether UK's \textit{leave} vote was caused by austerity-induced welfare reforms proposed by the UK government in 2010 \cite{fetzer2019did} has been calming down. Another interesting thing to note is the enormous drop of about 0.25 in cosine similarity for the word \textit{debacle} after June 23 2019, which has gained the most in terms of semantic relatedness to the word \textit{Brexit} in 2018. It is possible that this gain is related to the constant delays in the UK's attempts of leaving the EU). 

Some findings of the model are harder to explain. For example, according to the model, the talk about the \textit{renegotiating} in the context of \textit{Brexit} has not been very common in years 2015 and 2016 and than we can see a major rise of about 0.15 in cosine similarity in year 2017. On the other hand, an almost identical word \textit{renegotiation} has kept a very steady cosine similarity of about 0.72 throughout an entire five year period. We also do not have an explanation for a large drop in semantic relatedness in 2019 between words \textit{chequers} and \textit{Brexit}, and \textit{climate} and \textit{Brexit}. 

\subsection{Immigration news}
\label{sec:immigration}

Here we asses the performance of the proposed approach in a multilingual English-Slovenian setting. Since the main point of these experiments is to detect differences and similarities in relative semantic shift in two distinct languages, we first define English-Slovenian word pairs that arguably represent some of the most common aspects of the discourse about immigration \cite{martinez2000immigration,borjas1995economic,heckmann2016integration,cornelius2005immigration}. These English-Slovenian matching translations are \textit{crime-kriminal}, \textit{economy-gospodarstvo}, \textit{integration-integracija} and \textit{politics-politika}. We measure the cosine similarity between time specific vector representations of each word in the word pair and a time specific vector representation of a word \textit{immigration}. 

The results of the experiments are presented in Figure \ref{fig.3}. First thing one can note is that in most cases and in most years English and Slovenian parts of a word pair have a very similar semantic correlation to a word \textit{immigration}, which suggest that the discourse about immigration is quite similar in both countries. The similarity is most apparent for the word pair \textit{crime-kriminal} and to a slightly lesser extent for the word pair \textit{politics-politika}. On the other hand, not much similarity in relation to a word \textit{immigration} can be observed for an English and Slovenian words for economy. This could be partially explained with the fact that Slovenia is usually not a final destination for modern day immigrants (who therefore do not have any economical impact on the country) and serves more as a transitional country \cite{garb2018coping}, therefore immigration is less likely to be discussed from the economical perspective.  

Figure \ref{fig.3} also shows some interesting language specific yearly meaning shifts. The first one is the rise in semantic relatedness between the word \textit{immigration} and the English word \textit{politics} in 2016. This could perhaps be related to the Brexit referendum which occurred in the middle of the year 2016 and the topic of \textit{immigration} was discussed by politicians from both sides of the political spectrum extensively in the referendum campaign. 

Another interesting yet currently unexplainable yearly shift concerns Slovenian and English words for \textit{integration} in 2019. While there is a distinct fall in semantic relatedness between words \textit{integration} and \textit{immigration}, we can on the other hand observe a distinct rise in semantic relatedness between words \textit{integracija} and \textit{immigration}.

\section{Conclusion}
\label{sec:conclusion}

We presented a research on how contextual embeddings can be leveraged for the task of diachronic semantic shift detection. A new method that uses BERT embeddings for creating time specific word representations was proposed and we showcase the performance of the new approach on three distinct corpora, LiverpoolFC, Brexit news and Immigration news.

The proposed method shows comparable performance to the state-of-the-art on the LiverpoolFC corpus, even though domain adaptation was performed only on the corpus itself and no additional resources (as was the case in the study by \newcite{del2019short}) were required. This shows that the semantic knowledge that BERT model acquired during its pretraining phase can be successfully transferred into domain specific corpora. This is welcome from the stand point of reduced time complexity (since training BERT or most other embedding models from scratch is very time consuming) and it also makes our proposed method appropriate for detecting meaning shifts in domains for which large corpora are not available. 

Experiments on the Brexit news corpus are also encouraging, since detected relative semantic shifts are somewhat in line with the occurrence of different events which affected the news reporting about Brexit in different time periods. Same could be said for the multi-lingual experiments conducted on the English-Slovenian Immigration news corpus, which is to our knowledge the first attempt to compare parallel meaning shifts in two different languages, and opens new paths for multilingual news analysis.

On the other hand, a lot of further work still needs to be done. While the results on the Brexit and Immigration news corpora are encouraging, a more thorough evaluation of the approach would be needed. This could either be done in comparison to a qualitative discourse analysis or by a quantitative manual evaluation, in which changes detected by the proposed method would be compared to changes identified by human experts with domain knowledge, similar as in \newcite{del2019short}. 

The method itself could also be refined or improved in some aspects. While we demonstrated that averaging embeddings of word occurrences in order to get time specific word representations works, we did not experiment with other grouping techniques, such as taking median word representation instead of an average or by using weighted averages. Another option would also be to further develop clustering aggregation techniques, similar as in \newcite{giulianellilexical}. While these methods are far more computationally demanding and less scalable than averaging, they do have an advantage of better interpretability, since clustering of word usages into a set of distinct clusters resembles the manual approach of choosing the word's contextual meaning from a set of predefined meanings. 

\section{Acknowledgements}
This paper is supported by European Union’s Horizon 2020 research and  innovation programme under grant agreement No. 825153, project EMBEDDIA (Cross-Lingual Embeddings for Less-Represented Languages in European News Media). The authors acknowledge also the financial support from the Slovenian Research Agency for research core funding for the programme Knowledge Technologies (No. P2-0103) and the project TermFrame - Terminology and Knowledge Frames across Languages (No. J6-9372). We would also like to thank Lucija Luetic for useful discussions.

\section{Bibliographical References}
\label{reference}

\bibliographystyle{lrec}
\bibliography{bibliography}

\end{document}